\documentclass[sigconf,anonymous=false]{acmart}
\makeatletter
\def\@ACM@checkaffil{
    \if@ACM@instpresent\else
    \ClassWarningNoLine{\@classname}{No institution present for an affiliation}%
    \fi
    \if@ACM@citypresent\else
    \ClassWarningNoLine{\@classname}{No city present for an affiliation}%
    \fi
    \if@ACM@countrypresent\else
        \ClassWarningNoLine{\@classname}{No country present for an affiliation}%
    \fi
}
\makeatother

\usepackage{courier}
\usepackage[inkscapelatex=false]{svg}
\usepackage{pdfpages}
\usepackage{amsmath}
\usepackage[ruled,vlined,algo2e]{algorithm2e}
\usepackage{algorithm}
\usepackage{algorithmic}
\usepackage{booktabs}
\usepackage{multicol}
\usepackage{multirow}
\usepackage{siunitx}
\usepackage{soul}
\usepackage{url}
\usepackage{graphicx}
\usepackage{amsthm}
\usepackage{booktabs}
\usepackage{comment}
\usepackage{subfigure}
\usepackage{multirow}
\usepackage{multicol}  
\usepackage{enumitem}
\usepackage{array}
\usepackage{float}
\usepackage{tikz}
\usepackage{balance}

\newcommand{\zzj}[1]{{\color{black} #1}}

\newcommand{\smix}{SpatialMixer\xspace}

\newcommand{\tmixs}{TemporalMixers\xspace}

\newcommand{\etal}{\emph{et al.}\xspace}

\newcommand{\ie}{\emph{i.e.,}\xspace}
\newcommand{\etc}{\emph{etc.}\xspace}
\AtBeginDocument{%
  \providecommand\BibTeX{{%
    \normalfont B\kern-0.5em{\scshape i\kern-0.25em b}\kern-0.8em\TeX}}}

\setcopyright{acmcopyright}
\copyrightyear{2018}
\acmYear{2022}
\acmDOI{XXXXXXX.XXXXXXX}

\author{Zijian Zhang}
\affiliation{%
  \institution{Jilin University}
  \institution{City University of Hong Kong}
  \country{}
  }
\email{zhangzj2114@mails.jlu.edu.cn}
\authornote{The three authors contributed equally to this research.}

\author{Ze Huang}
\affiliation{%
  \institution{City University of Hong Kong}
  }
\email{zehuang3-c@my.cityu.edu.hk}
\authornotemark[1]

\author{Zhiwei Hu}
\affiliation{%
  \institution{City University of Hong Kong}
  }
\email{zhiweihu4-c@my.cityu.edu.hk}
\authornotemark[1]

\author{Xiangyu Zhao}
\affiliation{%
  \institution{City University of Hong Kong}
  }
\email{xianzhao@cityu.edu.hk}
\authornote{Xiangyu Zhao is the corresponding author.}

\author{Wanyu Wang}
\affiliation{%
  \institution{City University of Hong Kong}
  }
\email{wanyuwang4-c@my.cityu.edu.hk}

\author{Zitao Liu}
\affiliation{%
  \institution{Guangdong Institute of Smart Education, Jinan University}
  }
\email{liuzitao@jnu.edu.cn}

\author{Junbo Zhang}
\affiliation{%
  \institution{JD Intelligent Cities Research}
  \institution{JD iCity, JD Technology}
  }
\email{msjunbozhang@outlook.com}

\author{S. Joe Qin}
\affiliation{%
  \institution{Lingnan University, Hong Kong}
  }
\email{joeqin@LN.edu.hk}

\author{Hongwei Zhao}
\affiliation{%
  \institution{Jilin University}
  }
\email{zhaohw@jlu.edu.cn}

\acmPrice{15.00}
\acmISBN{978-1-4503-XXXX-X/18/06}

\begin{document}
\copyrightyear{2023}
\acmYear{2023}
\setcopyright{acmlicensed}\acmConference[CIKM '23]{Proceedings of the 32nd
ACM International Conference on Information and Knowledge
Management}{October 21--25, 2023}{Birmingham, United Kingdom}
\acmBooktitle{Proceedings of the 32nd ACM International Conference on
Information and Knowledge Management (CIKM '23), October 21--25, 2023,
Birmingham, United Kingdom}
\acmPrice{15.00}
\acmDOI{10.1145/3583780.3614969}
\acmISBN{979-8-4007-0124-5/23/10}

\renewcommand{\shortauthors}{Zijian Zhang et al.}

\title{MLPST: MLP is All You Need for Spatio-Temporal Prediction}

\begin{abstract}
Traffic prediction is a typical spatio-temporal data mining task and has great significance to the public transportation system. Considering the demand for its grand application, we recognize key factors for an ideal spatio-temporal prediction method: efficient, lightweight, and effective. However, the current deep model-based spatio-temporal prediction solutions generally own intricate architectures with cumbersome optimization, which can hardly meet these expectations. To accomplish the above goals, we propose an intuitive and novel framework, MLPST, a pure multi-layer perceptron architecture for traffic prediction. Specifically, we first capture spatial relationships from both local and global receptive fields. Then, temporal dependencies in different intervals are comprehensively considered. Through compact and swift MLP processing, MLPST can well capture the spatial and temporal dependencies while requiring only linear computational complexity, as well as model parameters that are more than an order of magnitude lower than baselines. Extensive experiments validated the superior effectiveness and efficiency of MLPST against advanced baselines, and among models with optimal accuracy, MLPST achieves the best time and space efficiency.

\end{abstract}




\maketitle

\section{Introduction}

\begin{figure*}[t]
\centering
\includegraphics[scale=0.7]{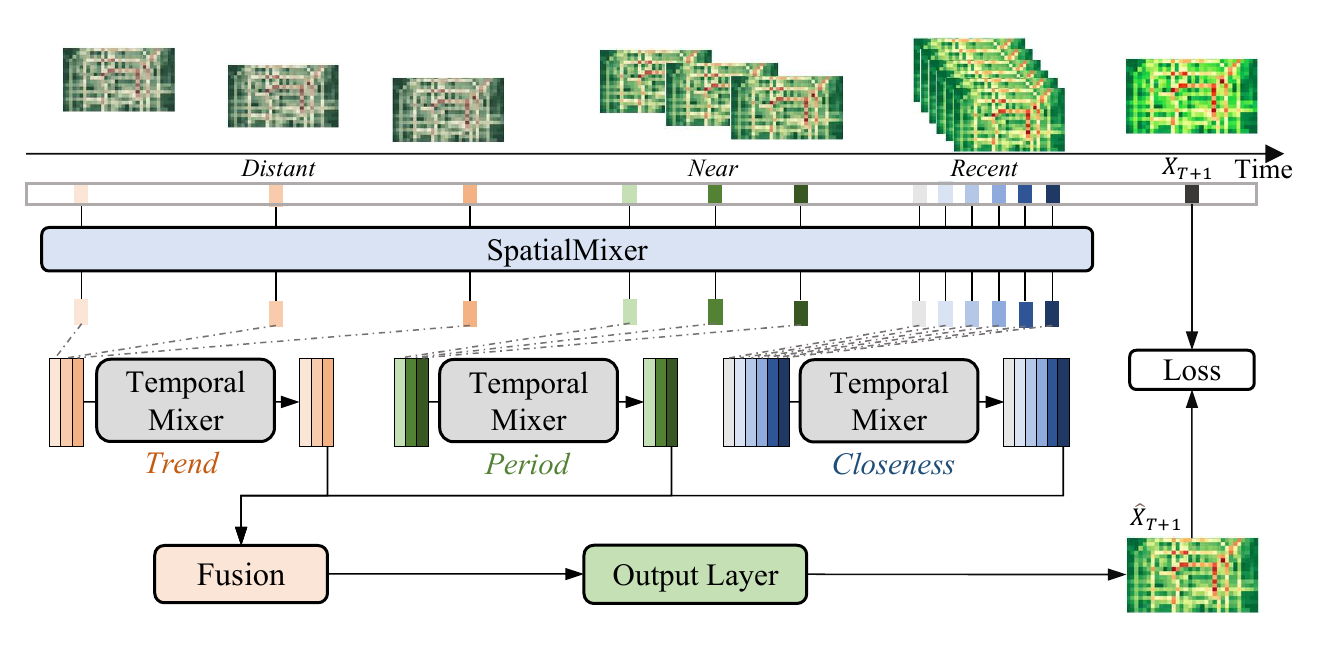}
\vspace{-5mm}
\caption{Framework overview for MLPST. } 
\label{Fig:1}
\end{figure*}
In recent years, urbanization has experienced substantial growth, leading to an exponential increase in the number of vehicles on the roads. The availability of accurate traffic prediction information is crucial for traffic management as it serves as a foundation for informed decision-making and can help drivers to navigate roads with reduced congestion and accidents \cite{min2011real, nagy2018survey, wang2021traffic, 10.1145/3580305.3599528, xu2016taxi}. 
The advancements in sensor technologies, mobile devices, and Global Positioning Systems (GPS) have enabled the collection of a wide range of Spatio-Temporal (ST) data in urban areas, which in turn, has facilitated the application of Spatio-Temporal Data Mining (STDM) in urban life 
\cite{wang2020deep, zhao2017modeling, zhao2018crime, zhao2017exploring, zhao2022multi, zhao2017incorporating}. 
STDM has widely fostered traffic prediction such as traffic flow analysis \cite{flow1, flow2, flow3}, travel time estimation \cite{time1, time2, zhang2023autostl}, and traffic demand prediction \cite{pan2012utilizing, demand1, demand2}.

Deep models have gained increasing popularity in recent years on STDM tasks. These models are known for their feature extraction and sequential modeling capacity, which enhances their effectiveness in extracting meaningful insights from ST data \cite{atluri2018spatio, wang2020deep}.
Spatial maps are often represented and processed as image-like matrices, which has led to the widespread application of Convolutional Neural Networks (CNNs) \cite{krizhevsky2012imagenet, meszlenyi2017resting, chen2018pcnn, tao2016deep, wang2017detecting} and Vision Transformers (ViT) \cite{liu2021swin, shi2020spatial, https://doi.org/10.48550/arxiv.2010.11929, han2021transformer}, which have been proven to be effective in capturing spatial features. On the other hand, for time series analysis, Recurrent Neural Networks (RNNs) \cite{ren2018deep, chung2014empirical, cho2014learning} and self-attention mechanisms \cite{Guo_Lin_Feng_Song_Wan_2019, huang2019dsanet, mohammadi2020transformer, garnot2020satellite} are commonly used to capture temporal correlations between different time slices. 

\zzj{
Although prosperous urbanization has greatly facilitated our lives, various pressing problems emerge and need to be resolved.
In consideration of the wide application of urban attribute prediction and the growing enormous quantity of data, we identify three crucial attributes for an effective spatio-temporal prediction method.
(i) \textbf{Efficient.} Swift training on new attributes and data.
(ii) \textbf{Lightweight.} Easy deployment and service with friendly storage requirements.
(iii) \textbf{Effective.} Well-capture the spatial and temporal dependency and guarantee precision of prediction.
However, current research can hardly meet the above ideal requirements.
Though achieve promising performance on spatio-temporal prediction, they highly depend on the intricate techniques ensembling \cite{lan2022dstagnn, wu2020connecting}, which takes overwhelming computational overhead and deployment cost \cite{shang2021discrete, dcrnn}.
}

To achieve the above targets, we propose an intuitive and novel framework, MLPST, which is made up of pure Multi-Layer Perceptrons (MLP). By using interleaved MLPs across certain dimensions, we can mix information from different parts of the input data. In this framework, we model spatial dependencies by patch processing the spatial map and conduct SpatialMixer between patches to learn global correlation within this spatial map, thus achieving a global receptive field. Besides, we define temporal dependencies to be three parts: closeness, period, and trend, so the temporal dependencies from different time intervals can be well modeled by the TemporalMixer. Benefiting from the simple and efficient MLP, our MLPST enjoys a rather low computational complexity, \ie $O(N)$. To achieve a fair comparison, we verify the performance of MLPST on real-world traffic flow datasets through an open-source platform LibCity \cite{wang2021libcity}. Extensive experiments prove the effectiveness of our framework. Results consistently show its superior performance compared to state-of-the-art methods. In a nutshell, the major contributions of our work can be summarized as follows: 

\begin{itemize}[leftmargin=*]
\item We propose an all-MLP framework for ST traffic prediction, MLPST. It consists of SpatialMixer and TemporalMixer, which are separately responsible for global spatial dependencies and temporal variations from different spans;

\item Our MLPST owns linear computational complexity and remarkably reduced parameters compared to existing methods, which indicates its promising potential for practical application;
\item We conduct extensive experiments to demonstrate the effectiveness of MLPST on two real-world datasets, which shows its superiority over existing baseline methods. We also present comprehensive analyzes to verify the validity of each component. 
\end{itemize}


\section{Framework}

In this section, we will present an all-MLP architecture, MLPST, which can solve STDM tasks in an effective way. We will start by introducing the overall architecture of our model and then elaborate on different modules in detail. Finally, we will provide the optimization process. 

\subsection{Preliminaries}
In this subsection, we first present the task definition and define the essential notations.

\noindent\textbf{\emph{Definition 1 (Traffic Flow).}} 
{
We split the city into ${H \times W}$ non-overlapping equal-sized grids based on longitude and latitude. 
Based on the traffic flow data in a certain period, we can calculate the traffic inflow and outflow of all grids. 
The traffic flow of all grids of a certain period $i$ is called a traffic flow grid map, noted as a matrix $\mathbf{X}_i$ $\in \mathbb{R}^{H \times W \times d}$, where $d=2$ represents traffic inflow and outflow.
}

\noindent\textbf{\emph{Definition 2 (Patch).}} 
A patch is defined as a $P \times P$ sized part of a grid map $\mathbf{X}_i$. Patches are non-overlapping parts of the original grid map. 
Thus, a grid map $\mathbf{X}_i$ is partitioned into $N_P=HW/P^2$ patches.

\noindent\textbf{\emph{Definition 3 (Spatio-Temporal Traffic Prediction).}} Given the historical observations $\{\mathbf{X}_i|i=1,2,\dots,T\}$, the goal of spatio-temporal traffic prediction is to predict the future traffic state $\mathbf{X}_{T+1}$, 
and $f(\cdot)$ stands for the spatio-temporal prediction model:
\begin{equation}
\label{equation1}
    \mathbf{X}_{T+1} = f([\mathbf{X}_1, \mathbf{X}_2, \dots, \mathbf{X}_T])
\end{equation}

\begin{figure}[t]
    \centering
    \includegraphics[scale=0.39]{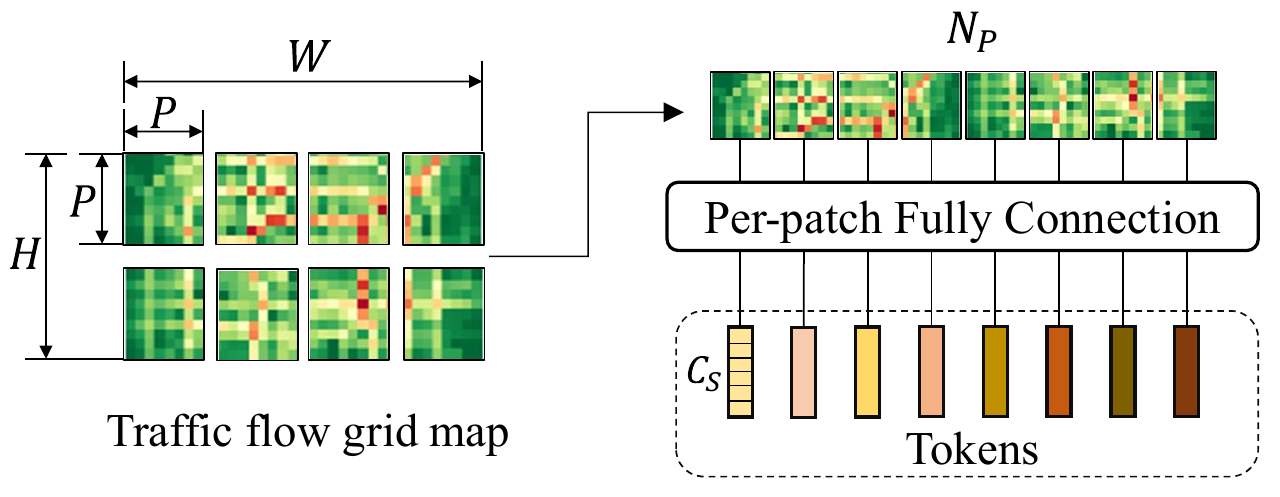}
    \caption{Patch processing. 
    }
    \label{Fig:2}
\end{figure}

\subsection{Framework Overview}
In this subsection, we present an overview of our framework. MLPST is a novel and efficient framework based on an all-MLP architecture, 
which mainly consists of several modules.
First, we propose two MLP-based modules, named SpatialMixer and TemporalMixer, to capture spatial and temporal dependency, respectively. Besides, we incorporate a fusion module to fuse the representation of different temporal dependencies. 
Finally, we employ an output layer to attain the prediction of the future traffic state. 

{
We introduce the pipeline as visualized in Figure \ref{Fig:1} from top to bottom.
Firstly we recognize the historical data sequence with three different kinds of temporal dependencies: \textit{trend}, \textit{period}, and \textit{closeness.} 
We address the spatial dependency by feeding \smix with the input feature of each time step, and obtain corresponding feature embeddings in a global view.
Then, we incorporate three \tmixs to capture the temporal relationship inside the three temporal dependencies, respectively. 
Finally, we aggregate the output embeddings of different temporal dependencies through a fusion module. The output layer further maps the fusion result to a $H \times W$ sized grid map to accomplish the single-step prediction.
}

\subsection{Detailed Modules}


\subsubsection{Input Layer} 
As aforementioned, the input layer is comprised of three fragments on the time axis denoting the observations from distant history, near history, and recent time. 
As illustrated in Figure \ref{Fig:1}, these fragments are used to model three types of temporal dependencies: \textit{trend}, \textit{period}, and \textit{closeness}, which are brought up based on sequencing observations with three kinds of intervals. Intuitively, the interval for trend series is the largest, which shows the development of data in a relatively long time, like seasonal changes. 
Period stands for the periodic variation of data in a fixed time period, for instance, traffic flow data often changes in one day cycle. 
Closeness means ST data is affected by the situation of recent time intervals. 
For example, if there is traffic congestion at 8 a.m., then the volume of vehicles at 9 a.m. is most likely to be high because of the closeness dependency.

Concretely Speaking, let $l_t$, $l_p$ and $l_c$ stand for the intervals of trend, period and closeness dependent sequences with the length of $t$, $p$ and $c$. 
Then the trend dependent sequence can be represented as a subset of all historical observations $[\mathbf{X}_{T - t \cdot l_t}, \dots,\mathbf{X}_{T - l_t}]$.
Likewise, the period and closeness dependent sequences can be defined as 
$[\mathbf{X}_{T - p \cdot l_p}, \dots,\mathbf{X}_{T - l_p}]$, and $[\mathbf{X}_{T - c \cdot l_c}, \dots,\mathbf{X}_{T - l_c}]$, 
where we assert $t+p+c=T$ so as to fix the input window. The usual situation is that $l_t > l_p > l_c$.
In this paper, we assign the temporal dependencies as follows: set the trend span $l_t$ to be one week which reveals the weekly trend; set $l_p$ to be a one-day period that describes the daily periodicity; set $l_c$ to be the unit interval to model the recent variation of the traffic flow. 


\begin{figure}[t]
    \centering
    \includegraphics[scale=0.31]{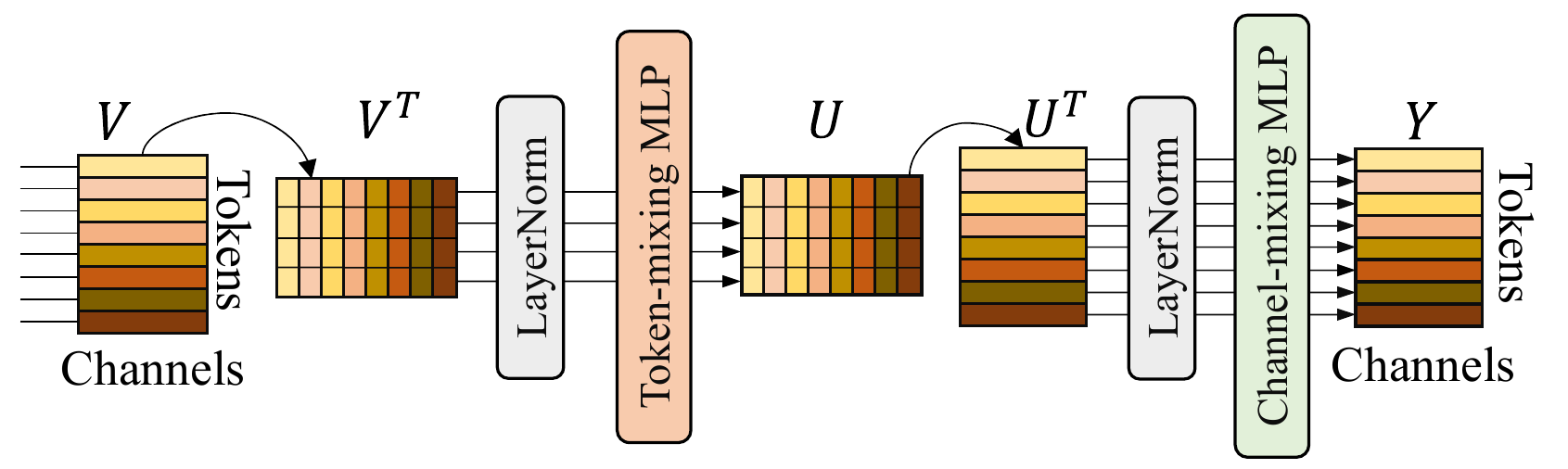}
    \caption{Visualization of all-MLP MixerLayer. }
    \label{Fig:3}
\end{figure} 

\subsubsection{SpatialMixer}
Modeling the region dependency comprehensively is not trivial, and we explore to tackle from both global and local perspectives. 
We propose an all-MLP SpatialMixer to model the spatial dependency in a linear complexity w.r.t. $N_P$. 
Specifically, in the SpatialMixer module, the spatial information of the input grid data is extracted and mixed to a certain-sized embedding $\mathbf{E}_{i}$. The dimension of $\mathbf{E}_{i}$ is related to the patch number $N_P$ and the channel number $C_S$, where \textit{channel} is defined as the hidden dimension of SpatialMixer for feature representation. SpatialMixer is comprised of the following blocks: Patch processing and All-MLP MixerLayers.

\textbf{\textit{Patch Processing}}. SpatialMixer starts with the patch processing step shown in Figure \ref{Fig:2}, including \textit{patch division} and \textit{per-patch fully connection}. Specifically, we first partition the $H \times W$ grid map into non-overlapping $P \times P$ sized patches. Each patch contains $P^2$ grids, so the flattened tensor belongs to $\mathbb{R}^{1 \times P^2}$. Assume the number of feature channels here is $C_S$, then we map the sequence of patches from $\mathbb{R}^{1 \times P^2}$ to $\mathbb{R}^{1 \times C_{S}}$ through per-patch fully connection. Every patch is mapped to a token, and all tokens are concatenated to be a two-dimension input matrix with the size of ${N_{P} \times C_{S}}$.

\textbf{\textit{All-MLP MixerLayer}}. We propose to capture the global correlations via an all-MLP MixerLayer \cite{tolstikhin2021mlp}.
We denote the output of patch processing as $\mathbf{V}$, which serves as the input matrix for MixerLayers. As demonstrated in Figure \ref{Fig:3},
MixerLayer performs LayerNorm, Token-mixing MLP, LayerNorm, and Channel-mixing MLP successively. 
The Token-mixing MLP is fomulated as: 
\begin{equation}
\label{Equation2}
\begin{split}
\mathbf{\mathbf{U}} =\mathbf{\mathbf{V}}^T+\mathbf{\mathbf{W}}_{2} \sigma\left(\mathbf{\mathbf{W}}_{1} \operatorname{LayerNorm}(\mathbf{\mathbf{V}}^T)+\mathbf{b}_1\right)+\mathbf{b}_2
\end{split}
\end{equation}
\noindent where $\sigma$ is a GELU activation, $\mathbf{W}_1,\mathbf{W}_2$, $\mathbf{b}_1$, and $\mathbf{b}_2$ are learnable weight matrices and bias.

\begin{figure}[t]
    \centering
    \includegraphics[scale=0.342]{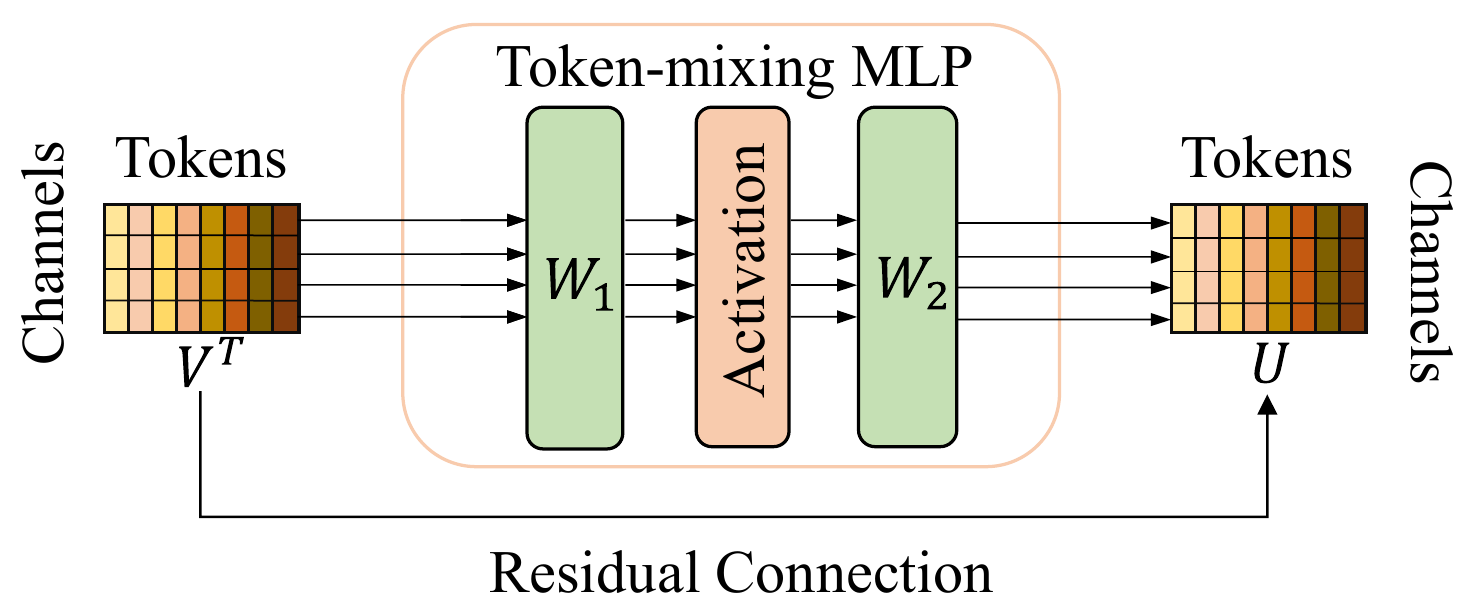}
    \caption{Illustration of Token-mixing MLP architecture.}
    \label{Fig:4}
\end{figure}

Token-mixing MLP is an MLP block that takes columns of matrix $\mathbf{V}$ as input, \ie the rows of the transposed matrix $\mathbf{V}^T$ as Figure \ref{Fig:4} illustrated. The MLP block maps $\mathbf{V}^T \in \mathbb{R}^{C_S \times N_P}$ to $\mathbf{U} \in \mathbb{R}^{C_S \times N_P}$.


The output of Token-mixing MLP $\mathbf{U}^T$ is fed to Channel-mixing MLP. As shown in Equation (\ref{Equation3}), Channel-mixing MLP maps $\mathbf{U}^T \in \mathbb{R}^{N_P \times C_S}$ to $\mathbf{Y} \in \mathbb{R}^{N_P \times C_S}$, which is the output of the MixerLayer.
\begin{equation}
\label{Equation3}
\begin{split}
\mathbf{\mathbf{Y}} =\mathbf{\mathbf{U}}^T+\mathbf{\mathbf{W}}_{4} \sigma\left(\mathbf{\mathbf{W}}_{3} \operatorname{LayerNorm}(\mathbf{\mathbf{U}}^T)+\mathbf{b}_3\right)+\mathbf{b}_4\end{split}
\end{equation}
where $\mathbf{W}_3,\mathbf{W}_4$, $\mathbf{b}_3$, and $\mathbf{b}_4$ are the learnable MLP parameters.

The fully-connected operations within Token-mixing MLPs and Channel-mixing MLPs 
enable the interaction between different spatial locations, \ie tokens, and channels. 
To conclude, MixerLayer maps $\mathbf{V}^T \in \mathbb{R}^{C_{S} \times N_{P}}$ to $\mathbf{Y} \in \mathbb{R}^{C_{S} \times N_{P}}$, which means the dimension of the input tensor does not change 
in MixerLayer. But at the same time, cross-location and per-location information can be captured with the composition of the Token-mixing and Channel-mixing. It is noteworthy that since tokens come from different locations on the original spatial map, then communication between tokens realizes the global mixing of spatial information, which effectively helps the feature extraction of spatial dependency. Besides, since $C_S$ is selected independently of $N_P$, the complexity of this part should be $O(N_P)$, unlike $O(N_P^2)$ of self-attention mechanisms.

After the processing of $N$ MixerLayers, for the convenience of following computation and interpretation, we formulate the output tensor to $\mathbf{E}_i \in \mathbb{R} ^ {1 \times N_{P} C_{S}}$ as the final output of SpatialMixer for time span $i$. Equation (\ref{Equation4}) represents the feature mapping process of SpatialMixer, where $\mathbf{E}_i$ denotes its output.
\begin{equation}
\label{Equation4}
\mathbf{E}_i = \operatorname{SpatialMixer}(\mathbf{X}_i),  \forall i = 1,2,...,T
\end{equation}

\subsubsection{TemporalMixer}
The time complexity is an inherent issue of temporal data mining, such as $O(T \cdot d^2)$ of RNN and $O(T^2 \cdot d)$ of self-attention, where $T$ represents the number of input time steps and $d$ is the embedding size.
Therefore, we propose to model the temporal dependencies via TemporalMixer consisting of a series of MixerLayers, and possesses a
reduced complexity of $O(T \cdot d)$.


As mentioned  before, 
we adopted three kinds of representations to model different temporal dependencies from the global view, \ie \textit{trend, period, closeness.} 
The embeddings $\mathbf{E}_i$ obtained by SpatialMixer of the same temporal dependency are concatenated as the input of a TemporalMixer.
Take the \textit{trend} dependent sequence as example, number of tokens is the sequence length $t$, and the concatenated input matrix for the TemporalMixer is $\mathbf{E}_t \in \mathbb{R}^{ t \times N_{P} C_{S}}$, as shown in Equation (\ref{Equation5}). 
Note that the internal architecture of TemporalMixer is identical with that of SpatialMixer, visualized in Figure \ref{Fig:4}.
Assume that the number of channels for TemporalMixer is $C_T$, which is the hidden dimension for its All-MLP MixerLayer. Then through Token-mixing and Channel-mixing, information in different parts of $\mathbf{E}_t$ interact with each other, \ie the trend dependency of certain locations can be captured.
\begin{equation}
\label{Equation5}
\mathbf{E}_t = \operatorname{Concat}(\mathbf{E}_i), \forall i = 1,2,\ldots,t
\end{equation}
Similarly, dimension of input tensor does not change through TemporalMixer. In other word, an All-MLP MixerLayer is utilized to capture temporal dependencies, which maps $\mathbf{E}_t \in \mathbb{R}^{ t \times N_{P} C_{S}}$ to $\widehat{\mathbf{E}}_t \in \mathbb{R}^{ t \times N_{P} C_{S}}$ as expressed by Equation (\ref{Equation6}). The complexity here is linear with respect to the number of tokens, which is $O(t \cdot d_T)$, where $d_T=N_P C_S$ is the embedding size for each time step.
\begin{equation}
\label{Equation6}
\widehat{\mathbf{E}}_t = \operatorname{TemporalMixer}(\mathbf{E}_t)
\end{equation}

Likewise, the same operations can be applied on the view of \textit{period} and \textit{closeness} dependencies, thus obtaining $\widehat{\mathbf{E}}_p$ and $\widehat{\mathbf{E}}_c$. The overall complexity of three parts is $O(T \cdot d_T)$ when $t+p+c=T$.

\subsubsection{Fusion}
In the Fusion module, we take the last time step of $\widehat{\mathbf{E}}_t$, $\widehat{\mathbf{E}}_p$ and $\widehat{\mathbf{E}}_c$ as mixing results of the three temporal dependencies. They are aggregated by assigning three corresponding learnable weights to different parts. Fusion module can be formulated by
:
\begin{equation}
\label{Equation7}
    \widehat{\mathbf{E}} = \mathbf{W}_t \circ \widehat{\mathbf{E}}_t\text{[t]} + \mathbf{W}_p \circ \widehat{\mathbf{E}}_p\text{[p]} + \mathbf{W}_c \circ \widehat{\mathbf{E}}_c\text{[c]}
\end{equation}
\noindent where $\circ$ means dot product, $[\cdot]$ denotes the index of the time dimension, and $\mathbf{W}_t,\mathbf{W}_p,\mathbf{W}_c$ are the learnable parameters that adjust the impact of trend, period, closeness. 

\subsubsection{Output Layer}
Output Layer is composed of a conventional MLP to map the embedding obtained by TemporalMixer to the size of the input grid, \ie Output Layer maps $\widehat{\mathbf{E}}$ to $\widehat{\mathbf{X}}_{T+1} \in \mathbb{R}^{H \times W}$, which achieves the single-step prediction and finishes the whole predictive learning pipeline.
\begin{equation}
\label{Equation8}
\widehat{\mathbf{X}}_{T+1} = \mathbf{W}_o \circ \widehat{\mathbf{E}}+\mathbf{b}_o
\end{equation}
\noindent where 
$\mathbf{W}_o$ and $\mathbf{b}_o$ are the weight and bias of output layer.

\subsection{Optimization}
In this section, we detail our loss function that is used for training. The loss function can be defined as:
\begin{small}
\begin{equation}
\label{Equation9}
    L(\theta) = \big(\sum_{i=1}^{M}|\widehat{\mathbf{X}}_{T+1}-\mathbf{X}_{T+1}|^q\big)^{1/q}
\end{equation}
\end{small}
\noindent where $\theta$ stands for the learnable parameters, 
and $M$ is the number of data records. 
$q$ indicates the regularization norm of loss function, such as $q=1$ of Mean Average Error (MAE) and $q=2$ of Root Mean Square Error (RMSE).


\subsubsection{Framework Complexity}
Since MLPST is an all-MLP architecture, the computational complexity of each individual module can be denoted as matrix multiplications. 

\textbf{\textit{SpatialMixer}} SpatialMixer starts feature extraction from patch processing with an output matrix $\mathbf{V} \in \mathbb{R}^{N_P \times C_S}$, and $\mathbf{V}$ serves as the input of all-MLP MixerLayers. Then the computational complexity of this MLP architecture can be denoted as several matrix multiplications, \ie $i \times h + h \times o$, where $i$ is the number of input units, $h$ is the number of hidden units, $o$ is the number of output units. As mentioned in the Framework section, MixerLayer consists of Token-mixing MLPs and Channel-mixing MLPs. The hidden unit of these MLPs is an expansion ratio that is independent of input units and output units, and the input units number equals the output units. So for Token-mixing MLPs, the complexity is $O(N_P)$ with $N_P$ being the number of input units. Likewise, the complexity for Channel-mixing is $O(C_S)$. Note that $C_S$ is chosen independent of the number of input tokens $N_P$, therefore, the complexity for a MixerLayer is $O(N_P)$.

\textbf{\textit{TemporalMixer}}
Structure of MixerLayers within TemporalMixer is identical with that of SpatialMixer. Similarly, for TemporalMixers that process different temporal dependencies, we can obtain each of their complexity. As for trend dependency, the input matrix is $\mathbf{E}_t \in \mathbb{R}^{t \times d_T}$. The number of input units for Token-mixing MLPs is the input sequence length $t$, so the complexity is $O(t)$ accordingly. Likewise, we have $O(d_T)$ for Channel-mixing MLPs. So the complexity of trend-dependent TemporalMixer is $O(t \cdot d_T)$. Similarly, we can get $O(p \cdot d_T)$ for period dependency and $O(c \cdot d_T)$ for closeness dependency. The add-up result of three parts should be $O(T \cdot d_T)$ when we assert $t+p+c=T$.

Based on the above deduction, the overall computational complexity for MLPST is $O(N_P \cdot T \cdot d_T)$.

\begin{figure}[t]
    \centering
    \includegraphics[scale=0.3]{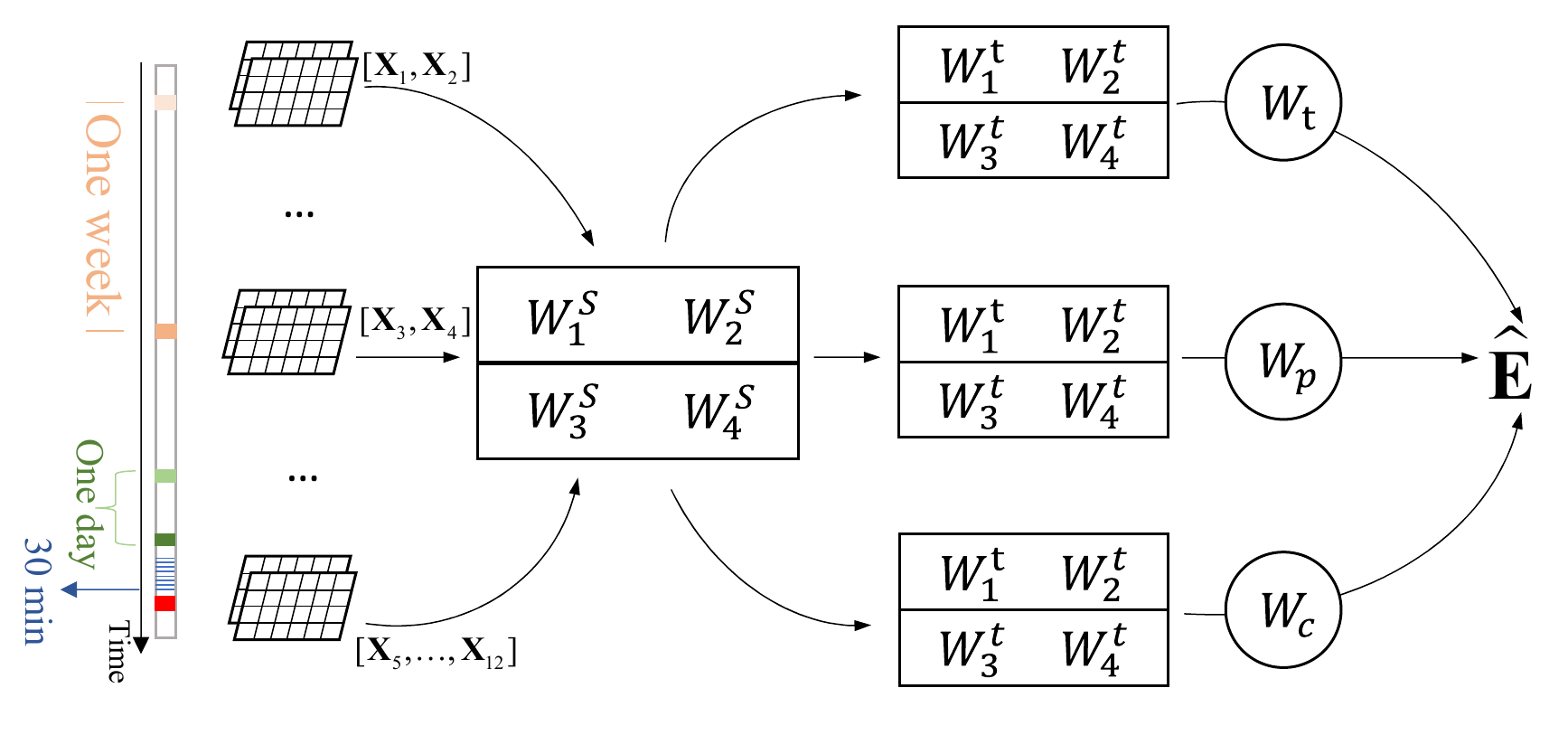}
    \caption{Traffic flow prediction.}
    \label{fig:pre}
\end{figure}

\subsection{Inference}
We present the detailed inference procedure and a toy example for traffic flow prediction in an example.
Specifically, we aim to make the single-step prediction with twelve input time steps of the traffic flow grid map $\{\mathbf{X}_i|i=1,2,\dots,12\}$, $\mathbf{X}_i \in \mathbb{R}^{H \times W \times 2}$, denoting the grid map size $H \times W$ and the feature dimension $2$ which means inflow and outflow respectively. Furthermore, we divide $[\mathbf{X}_1, \mathbf{X}_2, \dots, \mathbf{X}_{12}]$ into $[[\mathbf{X}_1, \mathbf{X}_2]$, $[\mathbf{X}_3, \mathbf{X}_4]$, $[\mathbf{X}_5,\dots,\mathbf{X}_{12}]]$, which denotes three types of temporal dependent sequences $[[trend],$
$[period],[closeness]]$. The learning process is performed by Token-mixing MLPs and Channel-mixing MLPs as pre-defined in Equation (\ref{Equation Token}) and (\ref{Equation Channel}):
\begin{equation}
\label{Equation Token}
    \mathbf{\mathbf{U}} =\mathbf{\mathbf{V}}^T+\mathbf{\mathbf{W}}_{2} \sigma\left(\mathbf{\mathbf{W}}_{1} \operatorname{LayerNorm}(\mathbf{\mathbf{V}}^T)\right)
\end{equation}
\begin{equation}
\label{Equation Channel}
    \mathbf{\mathbf{Y}} =\mathbf{\mathbf{U}}^T+\mathbf{\mathbf{W}}_{4} \sigma\left(\mathbf{\mathbf{W}}_{3} \operatorname{LayerNorm}(\mathbf{\mathbf{U}}^T)\right)
\end{equation}
\noindent where $V_T$ is a input matrix for MixerLayers. For SpatialMixer, we assume that  $\mathbf{W}_1^S,\mathbf{W}_2^S,\mathbf{W}_3^S,\mathbf{W}_4^S$ are the well learned weights. For TemporalMixer, we assume that $\mathbf{W}_1^t,\mathbf{W}_2^t,\mathbf{W}_3^t,\mathbf{W}_4^t$ are the learned weights for trend dependency, and similarly we have  $\mathbf{W}_1^p,\mathbf{W}_2^p,\mathbf{W}_3^p$, $\mathbf{W}_4^p$ and  $\mathbf{W}_1^c,\mathbf{W}_2^c,\mathbf{W}_3^c,\mathbf{W}_4^c$ for period dependency and closeness dependency respectively. At last, the traffic flow feature embedding of next time slot can be predicted as:
\begin{equation}
    \widehat{\mathbf{E}} = \mathbf{W}_t \circ \widehat{\mathbf{E}}_t\text{[t]} + \mathbf{W}_p \circ \widehat{\mathbf{E}}_p\text{[p]} + \mathbf{W}_c \circ \widehat{\mathbf{E}}_c\text{[c]}
\end{equation}
where $\mathbf{W}_t,\mathbf{W}_p,\mathbf{W}_c$ are another three learned weights for adjusting different temporal dependencies.
Figure \ref{fig:pre} represents how we make predictions with the given learned weights using Equation (\ref{Equation Token}) and Equation (\ref{Equation Channel}).
Finally, the output layer maps  $\widehat{\mathbf{E}}$ to the original input size to accomplish the single-step prediction. 

\section{Experiment}
In this section, we conduct extensive experiments on two real-world datasets to evaluate the effectiveness of MLPST.

\begin{table*}[t]
\centering
\caption{Overall performance comparison of all methods on two datasets.}
\label{Table:1}
\begin{tabular}{c||ccc||ccc||c}
\hline
\textbf{Datasets}          & \multicolumn{3}{c||}{\textbf{NYC Bike}} & \multicolumn{3}{c||}{\textbf{NYC Taxi}} &                                 \\ 
Metrics           & \multirow{2}{*}{MAE $\downarrow$}      & \multirow{2}{*}{RMSE $\downarrow$}    & \multirow{2}{*}{R2 $\uparrow$}       & \multirow{2}{*}{MAE $\downarrow$}      & \multirow{2}{*}{RMSE $\downarrow$}    & \multirow{2}{*}{R2 $\uparrow$}       &                                 \\
Methods                            &          &         &          &          &         &          & \multirow{-3}{*}{Parameters(K)} \\ \hline
AutoEncoder    & 5.57    & 8.27   & 0.687   & 22.45    & 65.77    &0.626     & \num{4.89e2}  \\
ResLSTM        & 3.44    & 10.72   & 0.472   & 19.74    & 59.86   & 0.715   & \num{1.74e4}  \\
Conv-GCN       & 5.19    & 7.71   & 0.728   & 11.80    & 20.58    & 0.966   & \num{3.48e3}  \\
Multi-STGCnet  & 1.41     & 4.95   & 0.885   & 9.31    & 42.13    & 0.844   & \num{1.81e4}  \\
ST-ResNet      & \underline{1.23}     & 3.37   & 0.946   & 8.60    &  63.80   & 0.631    & \num{1.80e3}  \\
FC-RNN         & 5.72     & 7.99   & 0.707   & 8.13    &  18.08   & 0.971   &   \num{0.69e2}   \\
Seq2Seq        & 4.93     & 6.83   & 0.786   & 7.78    & 17.74    & {0.972}   & \num{2.43e2}  \\
ACFM           & 1.26   & \underline{3.25}   & \underline{0.950}   & \underline{5.37}    &  \textbf{11.58}   & \underline{0.987}   & \num{3.39e2}\\
MTGNN   & 1.72 & 5.28 & 0.876 & 7.25 & 19.95 & 0.965 & \num{4.40e2}\\    
\hline
MLPST      & \textbf{1.20*}     &  \textbf{3.17*}   & \textbf{0.954*}   & \textbf{4.76*}    &  \underline{11.60}  & \textbf{0.988*}    & \textbf{0.60$\times 10^2$}                         \\ \hline
\end{tabular}
\\``\textbf{{\Large *}}'' indicates MLPST's statistically significant improvements (i.e., two-sided t-test with $p<0.05$) over the best baseline.
\end{table*}

\subsection{Experimental Setting}
\subsubsection{Datasets}



We conduct experiments on two real-world datasets from the New York City OpenData\footnote{https://opendata.cityofnewyork.us/}, \ie  NYCBike and NYCTaxi, which contain records of taxi and bike orders in New York City.

\textbf{NYCBike}: The NYCBike dataset contains bicycle trajectories collected from the NYC CitiBike system. The transaction records between July 1st, 2016, and August 31st, 2016 are selected. The following information is contained: bike pick-up station, bike drop-off station, bike pick-up time, bike drop-off time, and trip duration.

\textbf{NYCTaxi}: The NYCTaxi dataset contains tracks of different types of cabs collected by GPS for New York City from 2009 to 2020. We pick data from January 1st, 2015, to March 31st, 2015.
It provides properties including start and arrival time, geological information of start and end points, and trip distance.


\subsubsection{Evaluation Metrics}
To evaluate the effectiveness of our framework on traffic flow prediction, we adopt some commonly used value-based metrics, including Mean Absolute Error (MAE), Root Mean Squared Error (RMSE), and Coefficient of Determination ($R^{2}$). 


\subsubsection{Implementation}
We implement MLPST based on a public library, LibCity~\cite{wang2021libcity}, which contains several of the most representative traffic flow prediction models.

For the SpatialMixer part, 
(a) \textit{Per-patch Fully Connection}: first, we divide the city region into $10 \times 20$ disjoint grids of equal size. Then we set the patch size $P \times P$ to be $2 \times 2$.
According to the input data size, the number of channels $C_{S}$ is set to 20, which is the mapped dimension size of per-patch FC.
(b) \textit{MixerLayer}: 
we set the initial depth of MixerLayer to 8. As mentioned before, the number of patches $N_{P}$ can be expressed as $HW/P^2=50$, and the number of channels $C_{S}$ is known to be $20$. The hidden parameter $\text{expansion}$ is set to be 8, which is the number of hidden units in MLPs that are used in Token-Mixing and Channel-Mixing.

For the TemporalMixer part, 
(a) \textit{Input layer setting}: it is made up of $N$ MixerLayers, where the mixer depth $N$ is 8, and the number of channels for TemporalMixer $C_T$ is set to 20. We divide the input twelve time steps data into closeness, period, and trend. The initial values are 8, 2, and 2, respectively.

For other settings, 
(a) \textit{Optimization}: 
MLPST is implemented with Pytorch. The training process uses the Adam optimizer with the learning rate of \num{1e-3} and batch size of 64. We use cross-validation and early stop strategy. We utilize both MAE and RMSE error in our objective function, \ie $q=1,2$.
(b) We use Ray tune to optimize hyperparameters for our model. Most of the experiments for the baseline are performed with the released code in LibCity. To make the baseline results comparable, we adjust the number of input steps and do parameter tuning for all baselines to maintain uniformity. Each baseline model is trained five times to obtain an average result. 

\subsubsection{Baseline}
We compare the proposed MLPST with the following state-of-the-art baseline models: 

\begin{itemize}[leftmargin=*]
  \item\textbf{AutoEncoder}~\cite{hinton2006reducing}: it uses an encoder to learn the embedding vector from the data and then uses a decoder to predict future traffic conditions.
  \item\textbf{ResLSTM}~\cite{zhang2019deep}: it is a deep learning architecture merging the residual network, graph convolutional network, and long short-term memory to predict urban rail short-term passenger flow.
  \item\textbf{Conv-GCN}~\cite{zhang2020multi}: it combines a multi-graph convolutional network to address multiple spatio-temporal variations (\ie recent, daily, and weekly) separately. Then it uses a 3D convolutional neural network to integrate the inflow and outflow information deeply. 
  \item\textbf{Multi-STGCnet}~\cite{ye2020multi}: it contains three long short-term memory-based modules as temporal components and three spatial matrices as spatial components for extracting spatial associations of target sites.
  \item\textbf{ST-ResNet}~\cite{zhang2017deep}: it uses a residual neural network framework to model temporal closeness, period, and trend properties of crowd traffic, which is widely used in grid-based traffic prediction tasks.
  \item\textbf{FC-RNN}~\cite{wang2021libcity}: a swift deep learning model consisting of a fully connected layer and recurrent neural network for spatio-temporal prediction.
  \item\textbf{Seq2Seq}~\cite{sutskever2014sequence}: it is an encoder-decoder architecture based on gated cyclic units to predict the traffic state.
  \item\textbf{ACFM}~\cite{liu2018attentive}: Attention Crowd Flow Machine (ACFM) is capable of inferring the evolution of crowd flows by learning dynamic representations of data with temporal variation through an attention mechanism.
  \item\textbf{MTGNN}~\cite{wu2020connecting}: MTGNN solves multivariate time series prediction based on graph neural networks.
\end{itemize}

\subsection{Overall Performance}

Table \ref{Table:1} shows the overall performance of MLPST and the baseline models on the datasets NYCBike and NYCTaxi.

\begin{itemize}[leftmargin=*]
  \item As observed, the results of AutoEncoder on both datasets indicate that simply compressing and reconstructing the input data cannot extract useful features. 
  \item In contrast, complex deep learning models such as ResLSTM, Conv-GCN, and Multi-STGCnet get better results. However, there are differences in their performance results on the two datasets. These three models employ GCN to model the spatial dependencies between target stations. It shows that capturing the spatial dependencies within the whole network is essential. Both ResLSTM and Multi-STGCnet use LSTM to learn temporal correlations. So extracting features in the temporal dimension allows for better short-term passenger flow prediction. 
  From both spatial and temporal perspectives, we use modules composed of MLP-Mixer architectures in our framework MLPST, instead of complex deep learning structures like the three methods above. It is noteworthy that we obtain better results with our framework.
  \item ST-ResNet and Conv-GCN use different methods to model temporal closeness, period, and trend. ST-ResNet designs residual convolution units for each division and aggregates them dynamically. In our model, we similarly divide time steps into temporal closeness, period, and trend, but we apply MLP-mixer to each one and then fuse them by learnable weights. A reasonable division of input time steps improves traffic flow prediction accuracy.
  \item FC-RNN and Seq2seq process sequence data and learn temporal information. ACFM achieves spatial weight prediction using two ConvLSTMs units, where one LSTM learns an effective spatial-temporal feature representation.
  \item Our MLPST achieves consistently leading performance across all the metrics, which proves its advancing capability in modeling spatial and temporal dependencies.
  In addition to the superior prediction performance of MLPST, another advantage over the baseline is the relatively simple model structure and the small number of parameters. 
  {
  By sharing parameters between different layers, our MLPST owns the lowest amount of trainable parameters, \ie \num{0.60e2}, which is lower by more than one order of magnitude than baseline methods.
  }
  
\end{itemize}

To summarize, the above overall experimental performance demonstrates the effectiveness and efficiency of MLPST against representative baselines, which validates its superiority in spatio-temporal prediction tasks.
MLPST uses a simple all-MLP structure for traffic flow prediction in both the spatial and temporal dimensions. Moreover, we divide time steps in the temporal dimension. MLPST accomplishes complex tasks with a simple structure and performs well beyond the baseline.

\begin{table*}[!t]
\caption{Efficiency study of MLPST.}
\centering
\begin{tabular}{c||ccc||ccc}
\hline
\textbf{Dataset}                 & \multicolumn{3}{c||}{\textbf{NYCTaxi}} & \multicolumn{3}{c}{\textbf{NYCBike}}\\ 
Methods                 & MAE   & Train (s) & Infer (ms)  & MAE   & Train (s) & Infer (ms)  \\ \hline
AutoEncoder             & 22.45 & 1,573   & 47 & 5.57 & 1,443   & 62     \\
ResLSTM                 & 19.74 & 1,483   & 1,992 & 3.44 & 1,167   & 1,932     \\
Conv-GCN                & 11.80 & 297    & 160 & 5.19 & 157    & 150     \\
Multi-STGCnet           & 9.31  & 943    & 588 & 1.41  & 591    & 380     \\
ST-ResNet               & 8.60  & 4,208   & 1,613 & 1.23  & 6,845   & 1,781     \\
FC-RNN                  & 8.13  & 31     & 71 & 5.72  & 28     & 76     \\
Seq2Seq                 & 7.78  & 36     & 61 & 4.93  & 42     & 66     \\
ACFM                    & 5.37  & 4,775   & 2,319 & 1.26  & 5,843   & 2,891     \\ \hline
MLPST                   & \textbf{4.76}  & 132    & 38 & \textbf{1.20}  & 93    &39     \\ \hline
\end{tabular}
\label{Table:4}
\\Infer time is averaged for one batch (batch size = 64)
\end{table*}

\subsection{Efficiency Analysis}

In the context of swift urbanization, spatio-temporal model capacity, efficiency, and lightweight all play profound roles. This section surveys the efficiency of ST methods on the two datasets. To achieve a fair comparison, we conduct all experiments on one NVIDIA MX330 GPU. We follow the original settings for the baseline methods.

Table \ref{Table:4} shows the comparison results. 
For results on NYCTaxi, we observe that AutoEncoder, FC-RNN, and Seq2Seq obtain poor results for all three models despite their short training time and inference time due to their simple structures. The best-performing baseline ST-ResNet requires a significant amount of training time to improve its accuracy. 
Meanwhile, our MLPST achieves better results than ST-ResNet with only about $3\%$ of the training time. 
On NYCBike, we can observe that FC-RNN and Seq2Seq have very short training time and inference time because of their relatively simple structures, but these two models obtain inferior results to our MLPST. 
The best-performing baseline ACFM takes a lot of time on training and inference to improve its accuracy. We achieve superior results than ACFM, with only about $1\%$ of the training time and the inference time. 

The advanced performance, as well as training and inference efficiency of MLPST, can be ascribed to its simple structure and the incorporation of MLP-Mixer in both spatial and temporal dimensions. Therefore, MLPST obtains a better balance between time efficiency and performance.


\subsection{Ablation Study}
In this subsection, we will examine the effectiveness of each component of the model's architecture. 
MLPST mainly consists of two key components, \ie the SpatialMixer and the TemporalMixer. 
We remove our model's SpatialMixer and TemporalMixer architectures, respectively.
Then we obtain two variants of the original framework, \ie MLP-AT and MLP-SA. 
By comparing the performance of adjusted models above with MLPST, we can acquire insight into the effectiveness of these two modules.
In the following experiments, we implement these two variants of MLPST on the NYCtaxi dataset:
\begin{itemize}[leftmargin=*]
  \item\textbf{MLP-AT}: We remove the SpatialMixer, and keep the TemporalMixer unchanged for temporal feature learning.
  \item\textbf{MLP-SA}: We remove the TemporalMixer, and keep the SpatialMixer unchanged for spatial feature learning.
\end{itemize}

\begin{figure}[t]
    \centering
    \includegraphics[scale=0.63]{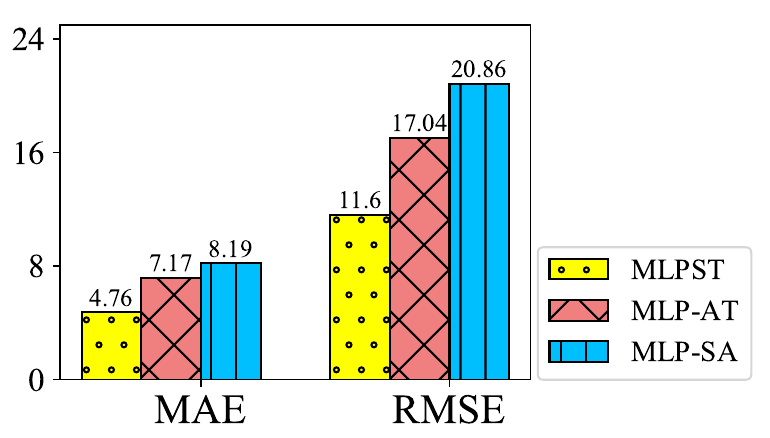}
    \caption{Results of ablation study on NYCTaxi dataset.}
    \label{Fig:5}
\end{figure}

As shown in Figure \ref{Fig:5}, the MAE obtained by MLP-AT is 7.17.
Model performance decreases by about 50$\%$ compared to MLPST, indicating that using the MLP-Mixer architecture in the spatial dimension effectively captures the spatial correlation between different grids.
MLP-SA worsens the model performance with the MAE increased by 3.43 because of removing TemporalMixer.
We observe that the model's prediction accuracy decreases sharply when the TemporalMixer is replaced. This result indicates that the MLP-Mixer architecture used by MLPST in the temporal dimension effectively captures valuable regions at each time step. Based on the experiments, it is obvious that each part of MLPST contributes as expected to the model. 

\subsection{Temporal Dependency Study}
In this section, we analyze the time step division in detail for the temporal dependency study to learn the influence of our MLPST on temporal dependency capture.

\begin{figure}[!t]
    \centering
    \includegraphics[scale=0.58]{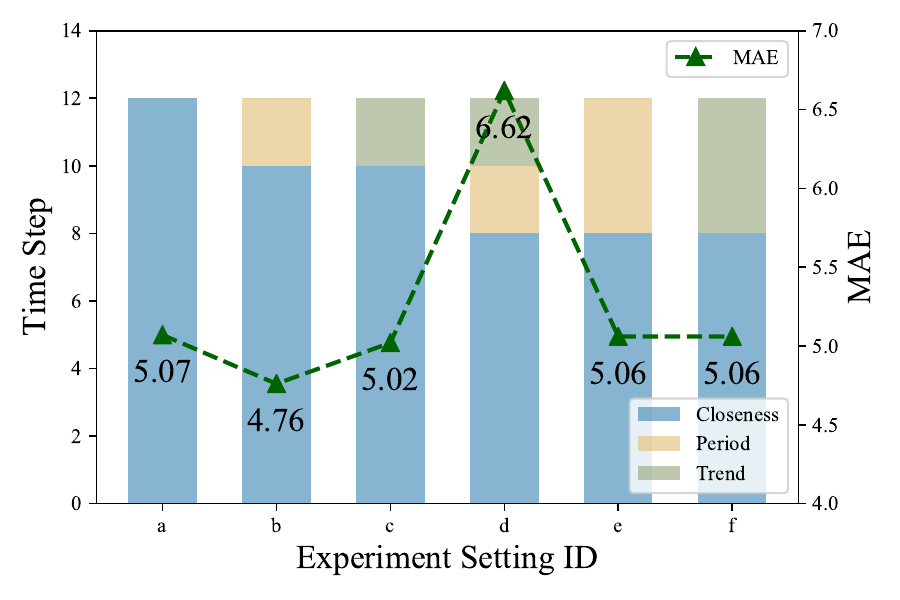}
    \caption{Impact of time step on NYCTaxi dataset.}
    \label{Figure:3}
\end{figure}

On the input side, we take a set of twelve time steps as input and output the predicted value of one time step. We select the critical time steps for modeling by dividing the twelve time steps into three groups: closeness, period, and trend. We fix the other parameters and make different adjustments to the division of the input data. For ease of representation, we use (closeness, period, trend) as a triplet. As shown in Figure \ref{Figure:3}, there are six groups of input data from a to f, which are (12, 0, 0), (10, 2, 0), (10, 0, 2), (8, 2, 2), (8, 4, 0), and (8, 0, 4). It does not make sense if the period is one or the trend is one because we need to do Token-mixing and Channel-mixing between embeddings in our model. 

From Figure \ref{Figure:3}, We can see that the input (10, 2, 0) gives slightly lower results compared to the input (12, 0, 0), and obtains the best result, which proves that the introduction of period benefits the model with the temporal dependency capture. However, because the results for input (8, 4, 0) dropped, it also indicates that introducing too long a period may not be useful or may be difficult to capture the information in it. The results for input (10, 0, 2) are worse than those for input (12, 0, 0) and (10, 2, 0). Finally, the results of input (8, 2, 2) and (8, 0, 4) do not work well, suggesting that the key closeness is still more helpful, which is in line with our perception of prediction in the spatial-temporal domain.

\subsection{Hyper-parameters Analysis}
Hyper-parameters in MLPST essentially influence the performance. This subsection will test the impact of 
the number of MixerLayer $N$, the number of channel in SpatialMixer $C_{S}$ and the number of channel in TemporalMixer $C_{T}$.

\begin{figure*}[t]
\centering
{\subfigure{\includegraphics[width=0.32\linewidth]{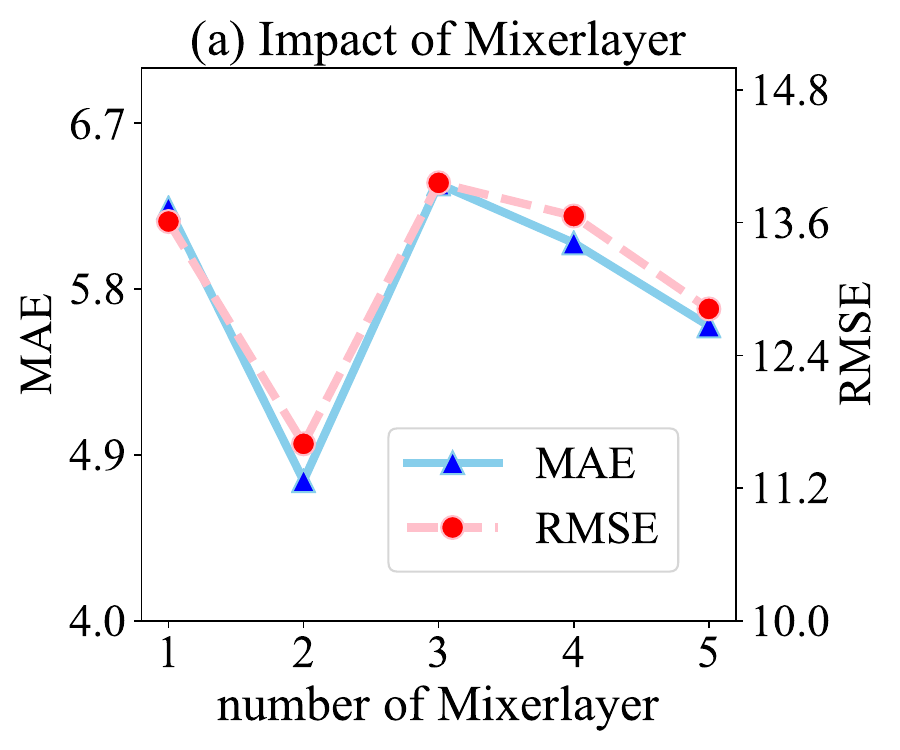}}}
{\subfigure{\includegraphics[width=0.32\linewidth]{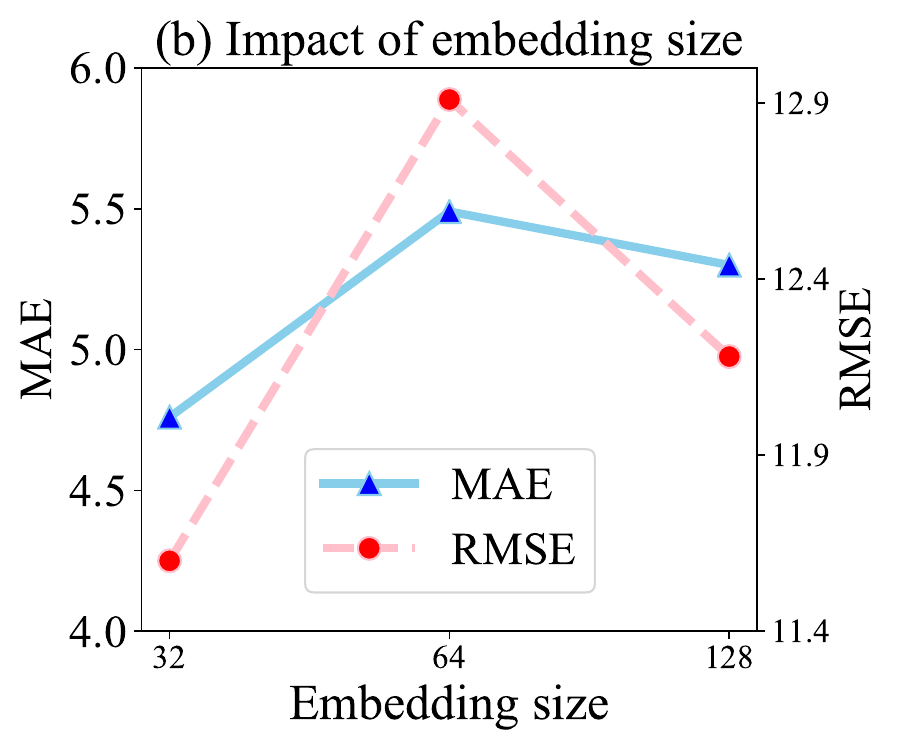}}}
{\subfigure{\includegraphics[width=0.32\linewidth]{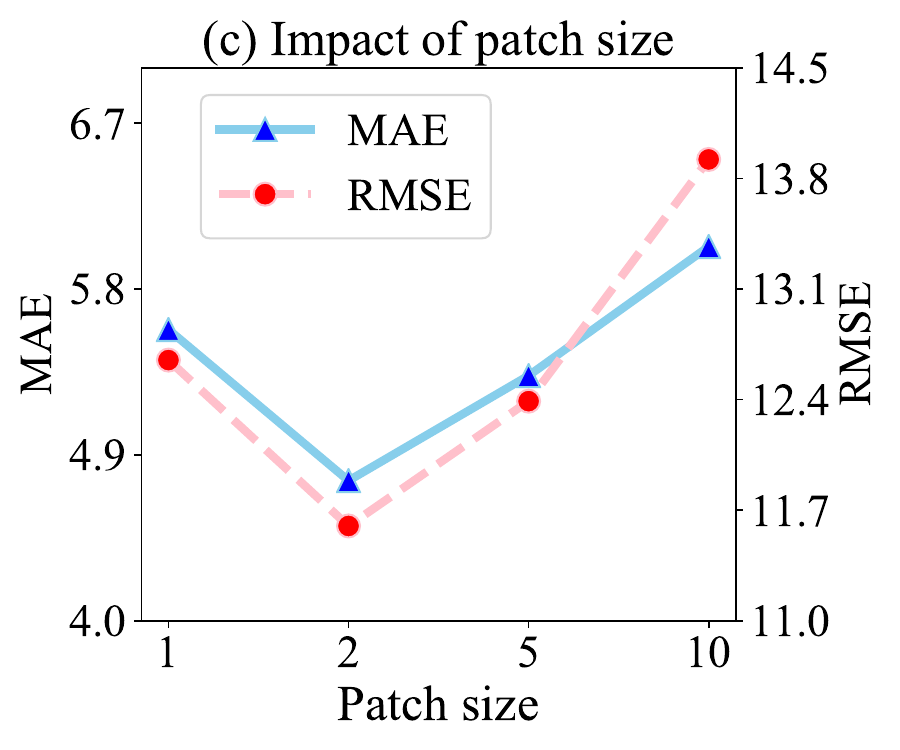}}}
\caption{Impact of Hyper-parameters on NYCTaxi dataset.} 
\label{Fig:6}
\end{figure*}

As we can see from Figure \ref{Fig:6} (a), the number of MixerLayer $N$ affects the results and time efficiency of the whole experiment. We find that the model performs the worst when $N=3$. When $N<6$, our model is relatively effective in terms of training time and exhibits outstanding performance. It proves MLPST's superior efficiency and efficacy in capturing spatio-temporal dependencies.
When $N>3$, We find the results fluctuate but get better as the $N$ continues to increase. The possible reason is that the architecture with deep MixerLayer often has an advantage in capturing global spatial-temporal dependencies.
Overall, as the MixerLayer $N$ increases, the model performance gets worse and then better.

For the number of channels in SpatialMixer $C_{S}$ and TemporalMixer $C_{T}$, we set them with equal values, \ie $C_{S}=C_{T}=C$. From Figure \ref{Fig:6} (b), the model performs best when $C=32$.

{
In Figure \ref{Fig:6} (c), we illustrate the performance of our MLPST with different patch sizes.
As aforementioned, we patch process the input spatio-temporal feature matrix and then process it with SpatialMixer and TemporalMixer.
From the results, we can observe that the best performance emerges when the patch size is 2. Given the input grid matrix with shape of $10\times 20$, it means 50 patches to be handled in parallel.
MLPST addresses the correlations among the 50 patches.
As the patch size rises to 5 and 10, it means the input matrix is divided into 4 and 2 patches, respectively.
Due to the sparse common information among the few patches, MLPST gets worse results.
What's more, we also test MLPST when the patch size is 1, \ie each grid serves as a patch and the input matrix is represented by  $10\times 20$ patches.
It is not inconceivable that MLPST gets lower performance with patch size of 2, because the patch is utilized to describe local spatial dependency, and taking each grid as a unit eliminates this locality.
}


\section{Related Work}

\noindent\textbf{MLP-Based Model.}
Since the emergence of CNN \cite{krizhevsky2012imagenet} and ViT \cite{vaswani2017attention}, they have been the most popular methods for computation vision. Later on, MLP-Mixer \cite{tolstikhin2021mlp} is proposed as an all-MLP framework for vision. ZHAO \etal \cite{zhao2021battle} argue that MLP-Mixer can achieve strong performance with small-sized models, but it can exhibit serious overfitting when the size of the model increases.
MLP-Mixer has been followed by various modified frameworks for better performance or lower complexity. 
ResMLP \cite{touvron2021resmlp} is another similar structure by Facebook consisting of Residual MLPs without normalization-based statistics. 
CycleMLP \cite{chen2021cyclemlp} proposed the idea of Cycle Fully-Connected Layers to enlarge the receptive fields 
for dense prediction tasks. 
gMLP \cite{tolstikhin2021mlp} proposed to use MLPs with gating and sMLP \cite{tang2022sparse} made adjustments on Token-mixing with sparse interaction, weight sharing and depth-wise convolution.

{
Unlike the MLP-Mixer methods in the related area, spatio-temporal data mining demands capturing both spatial and temporal relations.
In this paper, we propose a novel and effective MLP-based architecture to comprehensively model the global spatial information and the temporal patterns with multiple intervals.
To the best of the authors' knowledgement, MLPST is the first effort to address spatio-temporal prediction with an all-MLP framework.
}

\noindent\textbf{Spatio-temporal Data Mining.}
Most existing models for STDM tasks often adopt methods like encoders, CNN, graph convolutional networks (GCN), long short-term memory (LSTM), attention mechanisms, \etc
Seq2Seq \cite{sutskever2014sequence} and AutoEncoder \cite{hinton2006reducing} are designed based on an encoder-decoder architecture for spatio-temporal prediction.
STResNet \cite{zhang2017deep} is a CNN-based system designed for citywide ST prediction, and he proposed to make full use of temporal properties like closeness, period, and trend. 
ResLSTM \cite{zhang2019deep} and Multi-STGCnet \cite{ye2020multi} both adopt GCN for spatial feature extraction and LSTM for temporal dependency modeling. Conv-GCN \cite{zhang2020multi} also uses GCN to learn spatial correlation, but a 3D CNN is applied in temporal feature mining.
T-GCN \cite{zhao2019t} is a deep neural network method for traffic prediction, named temporal graph convolutional network, which combines GCN and GRU for spatial-temporal feature capturing.
ACFM \cite{liu2018attentive} stands for the attention crowd flow machine that can infer crowd flow and learn the dynamic representation of data through the attention mechanism. 

{
Spatio-temporal data mining methods aim to address the complex spatio-temporal dependency, which leads to the highly-complicated architecture and computational complexity.
In this paper, we prove that MLP-based architecture is enough for spatio-temporal characteristic capture without mechanisms with high complexity.
Our effective SpatialMixer and TemporalMixer can effectively model spatial and temporal dependency. It achieves state-of-the-art performance while attaining remarkable time and space efficiency.
}


\section{Conclusion}
In this paper, we propose a novel framework for solving STDM tasks, MLPST, which is an all-MLP architecture that is simple yet effective. To be specific, MLPST uses SpatialMixer and TemporalMixer targeting spatial and temporal view respectively for feature extraction.
On the one hand, SpatialMixer integrates interleaved MLPs on different spatial locations to model spatial dependency with global receptive. On the other hand, TemporalMixer captures the temporal correlations from multiple time intervals, which comprehensively considers temporal dependencies of different spans. 
We demonstrate the effectiveness of our framework by testing its performances on two real-world traffic flow datasets. The results show the superiority of our proposed framework against baselines, where MLPST attains optimal accuracy with the best efficiency.
Furthermore, MLPST is promising to be used in multiple types of STDM tasks in daily urban life, such as air quality prediction, urban energy consumption forecasting, \etc

\balance
\begin{acks}
This research was partially supported by APRC - CityU New Research Initiatives (No.9610565, Start-up Grant for New Faculty of City University of Hong Kong), CityU - HKIDS Early Career Research Grant (No.9360163), Hong Kong ITC Innovation and Technology Fund Midstream Research Programme for Universities Project (No.ITS/034/22MS), SIRG - CityU Strategic Interdisciplinary Research Grant (No.7020046, No.7020074), SRG-Fd - CityU Strategic Research Grant (No.7005894), Tencent (CCF-Tencent Open Fund, Tencent Rhino-Bird Focused Research Fund), Huawei (Huawei Innovation Research Program), Ant Group (CCF-Ant Research Fund, Ant Group Research Fund) and Kuaishou.
Zitao Liu is supported by National Key R\&D Program of China, under Grant No. 2022YFC3303600, and Key Laboratory of Smart Education of Guangdong Higher Education Institutes, Jinan University (2022LSYS003).
Junbo Zhang is funded by the National Natural Science Foundation of China (62172034), the Beijing Natural Science Foundation (4212021), and the Beijing Nova Program (Z201100006820053).
Partial financial support for this work from a Collaborative Research Fund by RGC of Hong Kong (Project No. C1143-20G), a grant from the Natural Science Foundation of China (U20A20189), and a Shenzhen-Hong Kong-Macau Science and Technology Project Category C (9240086).
Hongwei Zhao is funded by the Provincial Science and Technology Innovation Special Fund Project of Jilin Province, grant number 20190302026GX, Natural Science Foundation of Jilin Province, grant number 20200201037JC, and the Fundamental Research Funds for the Central Universities, JLU.

\end{acks}

\bibliographystyle{ACM-Reference-Format}
\bibliography{0Paper}


\end{document}